\documentclass[sigconf,natbib=false,nonacm=true]{acmart}

\usepackage{algorithm}
\usepackage{algorithmic}

\AtBeginDocument{%
  }

\RequirePackage[
  datamodel=acmdatamodel,
  style=acmnumeric,
  url=true,
  doi=false,
  isbn=false
]{biblatex}

\begin{document}

\title{Identifying the Geographic Foci of US Local News}

\author{Gangani Ariyarathne}
\orcid{0000-0003-4205-6574}
\affiliation{
  \institution{Department of Data Science}
  \institution{William \& Mary}
  \city{Williamsburg}
  \state{Virginia}
  \country{USA}
}
\email{gchewababarand@wm.edu}

\author{Isuru Ariyarathne}
\orcid{0000-0003-1811-7680}
\affiliation{
  \institution{Department of Data Science}
  \institution{William \& Mary}
  \city{Williamsburg}
  \state{Virginia}
  \country{USA}
}
\email{iahewababarand@wm.edu}

\author{Greatness Emmanuel-King}
\orcid{0000-0003-4205-6574}
\affiliation{
  \institution{Microsoft Corporation}
  \city{Redmond}
  \state{Washington}
  \country{USA}
}
\email{greatnesse@microsoft.com}

\author{Kate Lawal}
\orcid{0009-0003-0438-3571}
\affiliation{
  \institution{Mason School of Business}
  \institution{William \& Mary}
  \city{Williamsburg}
  \state{Virginia}
  \country{USA}
}
\email{kelawal@wm.edu}

\author{Alexander C. Nwala}
\orcid{0000-0003-3408-791X}
\affiliation{
  \institution{Department of Data Science}
  \institution{William \& Mary}
  \city{Williamsburg}
  \state{Virginia}
  \country{USA}
}
\email{acnwala@wm.edu}

\renewcommand{\shortauthors}{Gangani Ariyarathne, Isuru Ariyarathne, Greatness Emmanuel-King, Kate Lawal \& Alexander C. Nwala}

\begin{abstract}
Local journalism is vital in democratic societies where it informs people about local issues like, school board elections, small businesses, local health services, etc. But mounting economic pressures have made it increasingly difficult for local news stations to report these issues, underscoring the need to identify the salient geographical locations covered in local news (\textit{geo-foci}). In response, we propose a novel geo-foci model for labeling US local news articles with the geographic locations (i.e., the names of counties, cities, states, countries) central to their subject matter. First, we manually labeled US local news articles from all 50 states with four administrative division labels (\textit{local}, \textit{state}, \textit{national}, and \textit{international}) corresponding to their geo-foci, and \textit{none} for articles without a geographic focus. Second, we extracted and disambiguated geographic locations from them using Large Language Models (LLMs), since local news often contains ambiguous geographic entities (e.g., Paris, Texas vs. Paris, France). LLMs outperformed all eight geographic entity disambiguation methods we evaluated. Third, we engineered a rich set of spatial-semantic features capturing the prominence, frequency, and contextual positions of geographic entities. Using these features, we trained a classifier to accurately ($F_1$: 0.86) detect the geographic foci of US local news articles. Our model could be applied to assess shifts from local to national narratives, and more broadly, enable researchers to better study local media.
\end{abstract}


\maketitle

\section{Introduction}
US local news outlets report on the everyday realities of their communities, from school board decisions in Williamsburg, Virginia, to public health updates due to lead contamination in Milwaukee, Wisconsin. For over a decade, the loss of ad revenues, the consolidation of media ownership, and other factors have resulted in the \textit{nationalization of local news}~\cite{martin2019local} 
--- where local news stations prioritize state, national, or international issues over local ones. This raises the crucial question, ``does local journalism address the information needs of the communities they serve?'' To help address this question, we focus specifically on identifying the salient geographic regions covered by local news (\textit{geo-focus/foci}).

\begin{figure*}
\centerline{\includegraphics[width=\textwidth]{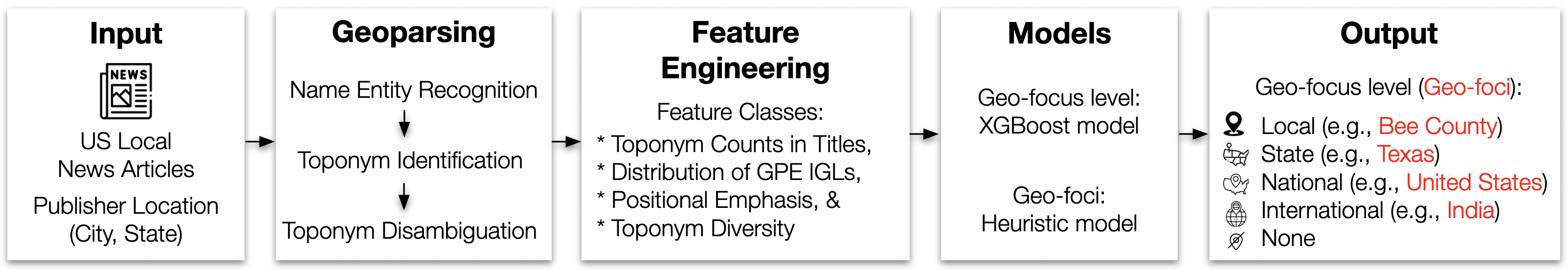}}
\caption{Overview of our geo-focus level classification  and geo-foci identification model (NLGF)}
\label{fig:design}
\Description{Overview of our geo-focus level classification  and geo-foci identification model (NLGF)}
\end{figure*}

Here, we define the \textit{geo-focus/foci} of a news article as the primary geographic area(s) central to its subject matter. This typically corresponds to the location(s) where key events occurred or the regions most affected by the story~\cite{d2014cliff}. We also define the \text{geo-focus level} for an article as one of these four administrative divisions central to its subject matter: \textit{local}, \textit{state}, \textit{national}, \textit{international}, and \textit{none} (for articles without a geographic focus). For example, the geo-focus of the \textit{Western Nebraska Observer} article titled, ``County [Kimball, Nebraska] Approves Motocross Track,'' is \textit{Kimball County, Nebraska}, and the geo-focus level is \textit{local}. Similarly, the geo-focus for the \textit{West Hawaii Today} article titled, ``Bangladesh air force jet crashes into school,'' is \textit{Bangladesh} and the geo-focus level is \textit{international}. We further expand on these in Section~\ref{sec:admin_div_details}. Our novel open-source News Lab Geo-Focus (NLGF) model (\textcolor{blue}{\href{https://github.com/wm-newslab/NLGF}{https://github.com/wm-newslab/NLGF}}) detects both the geo-foci and geo-focus level of US local news articles. Next, we briefly summarize its development.

First, we manually annotated a dataset of US local news articles from all US states with their respective geo-focus/foci and five \text{geo-focus levels}. Second, we extracted and disambiguated toponyms from the local news articles (\textit{geoparsing}~\cite{monteiro2016survey}). 
Specifically, we extracted toponyms of cities, counties, states, countries, etc., and disambiguated them by retrieving their respective latitude/longitude geo-coordinates along with the administrative divisions (county/state/country) the toponyms belong to. For toponym disambiguation, we evaluated six traditional geoparsers~\cite{halterman2023mordecai, grover2010use, geoparsepy_github, d2014cliff, gate_api, mendes2011dbpedia} and three Large Language Models (LLMs)~\cite{hurst2024gpt, touvron2023llama, abdin2024phi}. LLMs outperformed traditional geoparsers, especially when resolving ambiguous entities. After toponym disambiguation, we assigned each toponym an initial geo-focus level (IGL) based on the parent article's publisher's location. Next, we engineered a set of spatial-semantic features to train a classifier to accurately detect the geo-focus level ($F_1$: 0.89) of US local news articles and a heuristic model to identify their geo-focus/foci ($F_1$: 0.86).

Our contributions are as follows. First, our expert-annotated dataset is a valuable benchmark for future geo-focus research. Second, we adapted LLMs for toponym disambiguation and showed that they outperform traditional geoparsers. Third, we designed a set of spatial-semantic features that capture how geographic information is emphasized, distributed, and contextualized within articles. Finally, we combined these into an open-source classifier~\cite{gangani_nwala_nlgf}  that accurately determines the geo-foci (and geo-focus level) of US local news articles.

\section{Related Work}
\label{sec:related_work}

Geo-focus identification primarily involves two preceding steps: toponym recognition and toponym disambiguation.

\textbf{Toponym Recognition:} involves detecting geographic mentions (toponyms) such as names of cities, regions, etc., in text. This process traditionally relied on lookup, rules, and machine learning (ML)-based methods~\cite{monteiro2016survey}. More recently, statistical NLP methods like the Stanford Named Entity Recognition (NER), spaCy, and Apache OpenNLP, and transformer-based models like BERT and its derivatives~\cite{jehangir2023survey} are commonly used.

\textbf{Toponym Disambiguation:} Toponym recognition is insufficient due to ambiguous toponyms. Thus, toponym disambiguation involves accurately linking toponyms to their respective geo-coordinates. For example, disambiguating ``Paris'' to Paris, Texas, not Paris, France. It mainly involves three approaches~\cite{monteiro2016survey}: map-based using geometric algorithms or topological functions~\cite{bahgat2021toponym}, knowledge-based, which rely on gazetteers and semantic cues~\cite{alexopoulos2012optimizing}, and supervised methods that learn from annotated data~\cite{santos2015using}. Recently, LLMs with strong zero-shot capabilities have been shown to enhance toponym disambiguation~\cite{hu2024toponym}. In Section~\ref{sec:toponym-disambiguation}, we compared traditional geoparsers and LLMs for toponym disambiguation.

\textbf{Geographic Focus Identification}: aims to determine the geographic scope of an article. Cliff-Clavin~\cite{d2014cliff} used a frequency heuristic for this, assuming that the most frequently mentioned place is the geo-focus. Others incorporate ontologies~\cite{rodosthenous2018using}, rule-based classifiers~\cite{ahmad2021news}, and machine~\cite{imani2017focus}/deep~\cite{shingleton2024news}-learning models to improve geo-focus detection. Existing systems often lack generalizability in the face of region-specific vocabulary, variations in writing styles, and narrative structures. Transformer models perform well in many NLP tasks, but they struggle to capture fine-grained spatial-semantic clues needed for accurate geo-focus classification~\cite{shingleton2024news, tahmasebzadeh2023mm}.

\section{Methodology} \label{sec:methodology}

\begin{table*}[h]
\caption{Geoparsers vs. LLMs on the Toponym Disambiguation Task. Bold text marks best models for Geoparsers and LLMs.}
\begin{center}
\begin{tabular}{cc|cccccc|cccc}
\hline
\textbf{Category} & \textbf{Metric} & \textbf{Mordecai3} & \textbf{Edinburgh} & \textbf{Geoparsepy} & \textbf{Cliff} & \textbf{Gate Yodie} & \textbf{DBpedia} & \textbf{GPT-4o} & \textbf{LLaMA2-7b} & \textbf{Phi-3} \\
\hline
    GPE
     & Precision & \textbf{0.595} & 0.356 & 0.292 & 0.375 & 0.425 & 0.381 & \textbf{0.902} & 0.784 & 0.765 \\ 
     & Recall    & 0.611 & 0.727 & 0.824 & \textbf{0.857} & 0.607 & 0.640 & \textbf{1.000} & \textbf{1.000} & 0.949 \\ 
     & F1        & \textbf{0.603} & 0.478 & 0.431 & 0.522 & 0.500 & 0.478 & \textbf{0.948} & 0.879 & 0.847 \\ 
\hline
    LOC 
     & Precision & 0.321 & 0.282 & 0.316 & \textbf{0.475} & 0.308 & 0.409 & \textbf{0.700} & 0.450 & 0.495 \\ 
     & Recall    & 0.281 & 0.478 & 0.480 & 0.633 & 0.500 & \textbf{0.720} & \textbf{0.972} & 0.957 & 0.870 \\ 
     & F1        & 0.300 & 0.355 & 0.381 & \textbf{0.543} & 0.381 & 0.522 & \textbf{0.813} & 0.613 & 0.632 \\ 
\hline
    FAC 
     & Precision & 0.341 & 0.235 & 0.244 & 0.355 & \textbf{0.432} & 0.234 & \textbf{0.931} & 0.667 & 0.693 \\ 
     & Recall    & 0.583 & 0.320 & 0.500 & 0.355 & 0.731 & \textbf{0.733} & \textbf{1.000} & 0.957 & 0.813 \\ 
     & F1        & 0.431 & 0.271 & 0.328 & 0.355 & \textbf{0.543} & 0.355 & \textbf{0.964} & 0.785 & 0.748 \\ 
\hline
\end{tabular}
\label{tab:merged_evaluation}
\end{center}
\end{table*}

Here we explain our NLGF model for identifying the geo-foci and geo-focus levels of US local news articles. Fig.~\ref{fig:design} is a summary of the methodology.

\subsection{Data}
\label{sec:admin_div_details}
Our training dataset~\cite{gangani_nwala_nlgf} (Fig.~\ref{fig:design}: Input) consisted of 1,250 US local news articles evenly split across all five geo-focus levels. We built it by first randomly extracting local news articles (with publisher location) from the 3DLNews2~\cite{ariyarathne_nwala_3dlnews} dataset, for local news published in all US states. Next, two domain experts independently annotated each article with its geo-focus/foci and geo-focus level, using the annotation definitions in Section ~\ref{sec:annotation-definitions}. After annotation, we balanced the dataset to ensure equal representation across all geo-focus levels. We assessed annotation reliability by computing Inter-Rater Reliability (IRR) metrics across both the single-label geo-focus level and the multi-label geo-foci tasks. We observed strong annotation agreement for both tasks, confirming the robustness of the labels in the final dataset of 1,250 US local news articles. The geo-focus level annotations had a Cohen’s~$\kappa$ and Krippendorff’s~$\alpha$ of 0.83. The geo-foci annotations had a Krippendorff’s~$\alpha$ of 0.81.

\subsubsection{Annotation Definitions}
\label{sec:annotation-definitions}

\noindent
\\\textbf{Local:} Articles that focus on topics specific to a particular city, town, or county, such as local elections, school board decisions, or community events within a confined locality.

\noindent
\textbf{State:} Articles addressing topics relevant across multiple localities within the same US state, e.g., statewide elections.

\noindent
\textbf{National:} Articles covering issues of national importance or discussing topics involving other US states beyond the publisher’s home state, such as federal policies.

\noindent
\textbf{International:} Articles that focus on topics or events occurring outside the US, e.g., foreign affairs and global crises.

\noindent
\textbf{None:} Article with no clear geographic focus, instead discussing topics with broad appeal, such as scientific discoveries.

\subsection{Toponym Recognition}
We used spaCy (\textcolor{blue}{\url{https://spacy.io}}) for NER and extracted toponyms from these classes defined below: GPEs, LOCs, and FACs. For each toponym, we extracted its class, the sentence containing it, and the sentence's location (title/body).

\noindent
\textbf{Geo-Political Entities (GPEs):} includes geographic regions with political structures such as cities, states, and countries (e.g., \textit{Atlanta Georgia}, \textit{California}, \textit{Norway}).

\noindent
\textbf{Locations (LOCs):} refers to non-political locations (e.g., \textit{Rocky Mountains, Great Salt Lake}).

\noindent
\textbf{Facilities (FACs):} refers to buildings or structures that serve a specific purpose (e.g., \textit{The Pentagon, Statue of Liberty}).
\noindent

\subsection{Toponym Disambiguation} 
\label{sec:toponym-disambiguation}
We first tested six popular toponym disambiguation tools: Mordecai3~\cite{halterman2023mordecai}, Edinburgh Geoparser~\cite{grover2010use}, Geoparsepy~\cite{geoparsepy_github}, Cliff-Clavin~\cite{d2014cliff}, GATE YODIE~\cite{gate_api}, and DBpedia Spotlight~\cite{mendes2011dbpedia}. Each tool was tested using a gold-standard dataset of 300 manually disambiguated (with latitude/longitude) toponyms evenly split across each entity class (GPE, LOC, and FAC). The disambiguated toponyms were randomly selected from all US states (2 per state). The six tools performed inconsistently (Table~\ref{tab:merged_evaluation}), particularly for LOC and FACs, with low precision (\textless 0.5) and fair recall (\textless 0.75), motivating us to use LLMs for toponym disambiguation.

We evaluated these LLMs for toponym disambiguation: the proprietary GPT-4o~\cite{hurst2024gpt}, and the open-weighted LLaMA2-7b~\cite{abdin2024phi} and Phi-3~\cite{touvron2023llama} models. GPT-4o, significantly outperformed the rest across all classes, especially FACs (Table~\ref{tab:merged_evaluation}).

We did LLM toponym disambiguation as follows. First, each detected toponym (\textit{entity}), its entity class (\textit{entity\_class}: GPE, LOC, or FAC), its parent sentence (\textit{sentence}), and article publisher's location (\textit{city}, \textit{state}), were passed to all LLMs via this prompt: ``\textit{The following sentence is from a news media located in \{city\}, \{state\}. Disambiguate the \{entity\_class\} toponym entity `\{entity\}' in the sentence: \{sentence\}. Return both the coordinates (in decimal format) and the administrative level it refers to (county, state, or country). Format: latitude:\textless value\textgreater, longitude: \textless value\textgreater, administrative level (ADM):\textless county/state/country\textgreater.}'' 
The administrative level is needed to further disambiguate the toponym because a geo-coordinate alone is not sufficient to specify the geographic location the sentence refers to. For example, the geo-coordinate: (\textit{39.9612, -82.9988}) can point to either the city of Columbus, Ohio, or the state of Ohio, or even the United States as a whole. We parsed the LLM responses to extract latitude, longitude, and administrative levels. We used Shapely (\textcolor{blue}{\url{https://github.com/shapely/shapely}}) to verify that returned geo-coordinates are within polygons of US counties or states, or other countries.

After disambiguation, we assigned each disambiguated toponym an Initial Geo-focus Level (IGL) - \textit{local}, \textit{state}, \textit{national}, or \textit{international}. Recall that we define the \text{geo-focus level} for each article as one of these four administrative divisions central to its subject matter: \textit{local}, \textit{state}, \textit{national}, \textit{international}, and \textit{none} for articles without a geographic focus. The initial geo-focus level is similar to the geo-focus level, but unlike the geo-focus level which is associated with the entire article, the initial geo-focus level is only associated with the sentence containing a toponym. The disambiguated toponyms, along with their respective IGLs are used to generate features needed for NLGF to predict the geo-foci and geo-focus levels of US local news articles. We assigned IGLs to toponyms using the following logic that considers the target geographic audience of the news publisher. If the administrative level (ADM) of the disambiguated toponym is \textit{country}, and the country is USA, we assign the toponym the \textit{national} IGL; otherwise, \textit{international}. If the ADM is \textit{state}, and the state matches the news article's publisher's state, the toponym is assigned \textit{state} IGL; otherwise, \textit{national}. If the ADM is \textit{county}, and the county is within the article's publisher's state, the toponym is assigned \textit{local} IGL; otherwise \textit{national}.

\subsection{Geo-focus level and geo-foci identification}
\label{sec:geo_focus_classifier}

NLGF first labels a news article with one of the five geo-focus levels using an XGBoost geo-focus classifier. Next, it identifies the geographic foci (toponyms) using Alg.~\ref{alg:geo-foci}. 

\subsubsection{Feature Engineering}
\label{sec:feat_engineering}

We extracted 15 features across four feature classes to train the geo-focus level classifier. The features quantify the presence and prominence of geographic references in US local news articles for all geo-focus levels. The feature classes with actual feature names in parentheses are:

\noindent
\textbf{Toponym Counts in Titles}: The presence of toponyms in news titles signal spatial emphasis. We extracted the counts of toponyms in the titles of news articles for each geo-focus level (\textit{title\_topo\_cnt\_local}, \textit{title\_topo\_cnt\_state}, \textit{title\_topo\_cnt\_national}, \textit{title\_topo\_cnt\_intl}).

\noindent
\textbf{Distributions of GPE IGLs}: GPEs are more specific than LOCs or FACs, hence easier to disambiguate. We exploit this by counting the IGLs of GPE toponyms in articles by counting the GPEs for each IGL (\textit{local\_igl\_cnt}, \textit{state\_igl\_cnt}, \textit{national\_igl\_cnt}, and \textit{intl\_igl\_cnt}).

\noindent
\textbf{Positional Emphasis}: Toponyms located early in news articles could signal spatial emphasis. We define early as being one of the first five toponyms (leading toponym). We counted the leading toponyms for each IGL (\textit{leading\_topo\_local\_igl\_cnt}, \textit{leading\_topo\_state\_\\igl\_cnt}, \textit{leading\_topo\_national\_igl\_cnt}, and \textit{leading\_topo\_intl\_igl\_cnt}).

\noindent
\textbf{Toponym Diversity:} The presence of different geographic regions in news articles could indicate a broad geo-foci. We measured this by counting unique geographic identifiers for each IGL (\textit{uniq\_local\_igl}, \textit{uniq\_national\_igl}, and \textit{uniq\_intl\_igl}). We excluded the \textit{state} IGL since this represents only one location (publisher's state).  

\subsubsection{Geo-focus level/geo-foci identification}
\label{sec:geo_focus_level_foci_identification}

For geo-focus level classification, we used XGBoost, an ensemble of gradient-boosted decision trees, selected for its ability to model non-linear relationships. Hyperparameters, including learning rate, tree depth, number of estimators, and subsample ratio, were optimized to 0.2, 6, 25, and 0.9, respectively, via stratified 5-fold cross-validation with grid search, targeting macro-averaged $F_1$. Feature correlations were checked to ensure no pair exceeded 0.85.

\begin{algorithm}
\caption{Geo-Foci Identification}
\label{alg:geo-foci}
\begin{algorithmic}[1]
\STATE \textbf{Input:} Article $A$, geo-focus level $gl$, toponyms $T$ with IGLs
\STATE \textbf{Output:} Geo-foci $F$

\STATE $T' \leftarrow \{ t_i \in T \mid \text{IGL}(t_i) = gl \}$
\FOR{each $t_i \in T'$}
    \STATE $focus\_score(t_i) \leftarrow freq_{\text{title}}(t_i) + freq_{\text{article}}(t_i) + freq_{\text{lead}}(t_i) + freq_{\text{gpe}}(t_i)$
\ENDFOR
\STATE $S \leftarrow \sum_{t_i \in T'} focus\_score(t_i)$
\FOR{each $t_i \in T'$}
    \STATE $focus\_score(t_i) \leftarrow focus\_score(t_i) / S$
\ENDFOR
\STATE $F \leftarrow \{ t_i \in T' \mid focus\_score(t_i) > \alpha \}$
\STATE \textbf{Return} $F$
\end{algorithmic}
\end{algorithm}

Once NLGF predicts the geo-focus level $gl$ ($gl \in \{local, state, \\ national, international, none\}$) for an article, it identifies the article's geo-focus/foci by selecting top ranked toponyms (toponyms with \textit{focus\_score}  $> \alpha$) from the list $T$ of all toponyms used to predict the article's geo-focus level. This is done using Alg.~\ref{alg:geo-foci} summarized as follows. First, we discard toponyms from $T$ that have an IGL different from the identified $gl$ (Alg.~\ref{alg:geo-foci}, line 3). Recall that each toponym $t_i$ has a corresponding IGL (\textit{local}, \textit{state}, \textit{national}, \textit{international}). Second, we compute the \textit{focus\_score} for all remaining toponyms in $T$. We define the \textit{focus\_score} (Alg.~\ref{alg:geo-foci}, line 5) of a toponym $t_i$ ($t_i \in T$) as: frequency of $t_i$ in the article's title + total mentions of $t_i$ in the article + the number of times $t_i$ was a leading toponym + the number of times $t_i$ was recognized as a GPE. We normalized (Alg.~\ref{alg:geo-foci}, line 9) all \textit{focus\_score}s and selected toponyms with \textit{focus\_score} > $\alpha$ (Alg.~\ref{alg:geo-foci}, line 11), as the predicted geo-foci. We tuned $\alpha$ on a 3:1 train-test split by sweeping values from 0.05 to 0.5 (in increments of 0.05), and selected the value (0.25) that maximized the $F_1$.

To evaluate geo-foci predictions, where each article may contain zero, one, or multiple geo-foci, we used the scikit-learn library’s MultiLabelBinarizer to convert the set of true and predicted toponyms for each article into a binary indicator vector. Precision, recall, and $F_1$ were then computed using sample-based averaging, which evaluates performance at the article level by comparing each predicted set of foci to its corresponding ground truth set. We selected this evaluation strategy because geo-foci prediction is inherently a multi-label problem since articles may contain varying numbers of toponyms. Using sample-averaged precision, recall, and $F_1$ allows each article to be assessed as an independent decision instance, providing a fairer reflection of real-world performance than metrics aggregated across unevenly distributed toponyms.

Finally, we compared NLGF to two baselines, GPT-4o and Cliff-Clavin~\cite{d2014cliff}, a widely used benchmark for geographic focus identification. Cliff-Clavin was designed to identify geo-foci rather than geo-focus levels. However, the predicted geo-foci can be mapped to geo-focus levels using rule-based aggregation~\cite{gangani_nwala_nlgf}, enabling its use as a baseline for geo-focus level classification. GPT-4o was prompted with the article’s title, content, publisher location, and definitions of geo-focus and geo-focus levels (prompt details available in the NLGF repository~\cite{gangani_nwala_nlgf}).

A broader evaluation of additional baseline geo-focus identification systems was not feasible due to limitations in system availability and scope. For example, the GeoMantis system~\cite{rodosthenous2018using} that was reviewed in Section~\ref{sec:related_work} infers country-level geo-focus, and therefore cannot be evaluated on finer-grained geographic categories such as local, state, national, and none. Other approaches rely on proprietary or non-public implementations. Ahmad et al.~\cite{ahmad2021news} and Shingleton et al.~\cite{shingleton2024news} propose machine- and deep-learning models that are not public, and despite contacting the authors, we were unable to obtain access for replication. Imani et al.~\cite{imani2017focus} focus specifically on political news, making their model unsuitable for general news domains. Consequently, Cliff-Clavin was the only system from the literature that was accessible, domain-appropriate, and deployable for direct benchmarking.

\section{Results and Discussion}

\begin{figure}[H]
\centerline{\includegraphics[width=\linewidth]{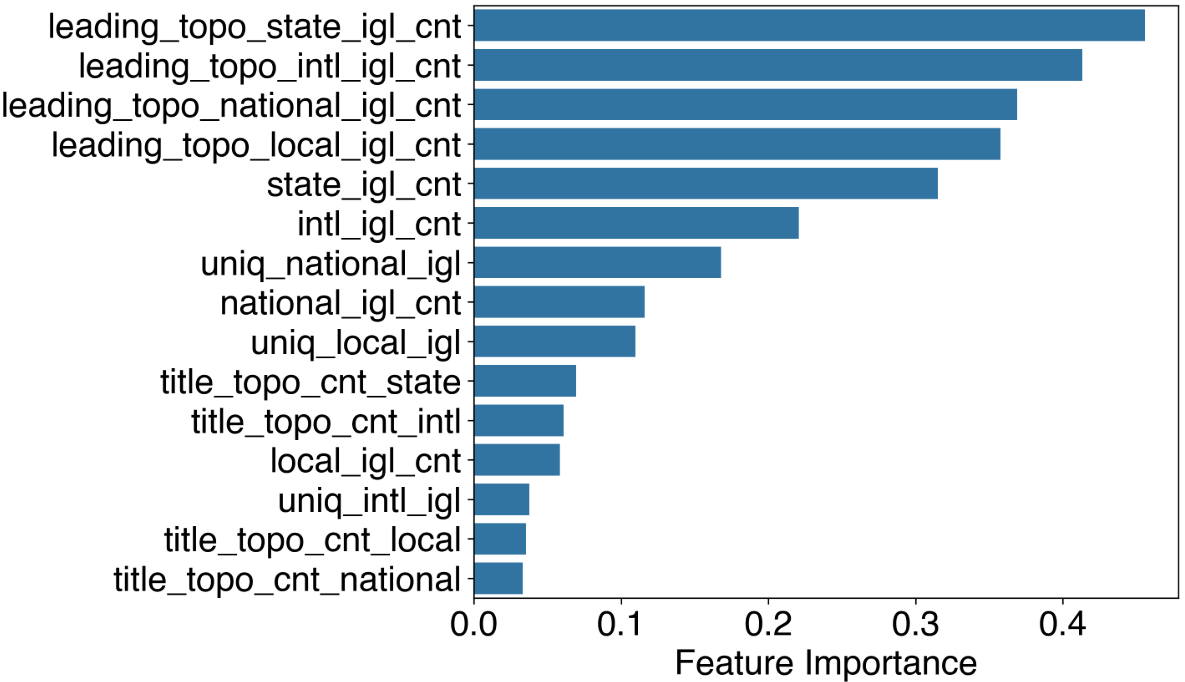}}
\caption{Feature importance of NLGF.}
\label{fig:feature_importance}
\Description{Feature importance of NLGF.}
\end{figure}

NLGF outperformed both baselines in the task of classifying the geo-focus levels and identifying the geo-foci of US local news articles. For geo-focus level classification (Table~\ref{tab:gpt_vs_xgb_cliff}), it achieved 0.89 macro-averaged precision, recall, and $F_1$, outperforming GPT-4o (precision: 0.76, recall: 0.75, $F_1$: 0.75) and Cliff-Clavin (precision: 0.67, recall: 0.64, $F_1$: 0.62). The feature importance plot (Fig.~\ref{fig:feature_importance}) further explains this performance gap. The most influential features are those capturing leading toponyms, including counts of leading international, national, state, and local toponyms. These features encode the geographic scope emphasized early in articles. 

\begin{table}[H]
\caption{NLGF vs. GPT-4o vs. Cliff-Clavin: Geo-focus level/Geo-foci prediction results.}
\begin{center}
\resizebox{\columnwidth}{!}{
\begin{tabular}{cccccccccc}
\hline
\textbf{Geo-focus} &
\multicolumn{3}{c}{\textbf{NLGF}} &
\multicolumn{3}{c}{\textbf{GPT-4o}} &
\multicolumn{3}{c}{\textbf{Cliff-Clavin}} \\
\cline{2-4} \cline{5-7} \cline{8-10}
\textbf{level} & Prec. & Rec. & $F_1$ & Prec. & Rec. & $F_1$ & Prec. & Rec. & $F_1$ \\
\hline
Local          & \textbf{0.85} & \textbf{0.88} & \textbf{0.87} & 0.62 & 0.74 & 0.67 & 0.43 & 0.58 & 0.49 \\
State          & \textbf{0.88} & \textbf{0.87} & \textbf{0.88} & 0.78 & 0.69 & 0.73 & 0.61 & 0.26 & 0.37 \\
National       & \textbf{0.83} & \textbf{0.89} & \textbf{0.86} & 0.81 & 0.60 & 0.69 & 0.64 & 0.77 & 0.70 \\
Intl.          & \textbf{0.93} & \textbf{0.96} & \textbf{0.94} & 0.89 & 0.92 & 0.90 & 0.70 & \textbf{0.96} & 0.81 \\
None           & \textbf{0.95} & \textbf{0.83} & \textbf{0.89} & 0.71 & 0.82 & 0.76 & \textbf{0.95} & 0.62 & 0.75 \\
\hline
Macro avg      & \textbf{0.89} & \textbf{0.89} & \textbf{0.89} & 0.76 & 0.75 & 0.75 & 0.67 & 0.64 & 0.62 \\
\hline
\hline
\textbf{Geo-foci}   & \textbf{0.86} & \textbf{0.89} & \textbf{0.86} & 0.65 & 0.69 & 0.66 & 0.29 & 0.64 & 0.37 \\
\hline
\end{tabular}
}
\label{tab:gpt_vs_xgb_cliff}
\end{center}
\end{table}

\begin{figure}
\centerline{\includegraphics[width=\linewidth]{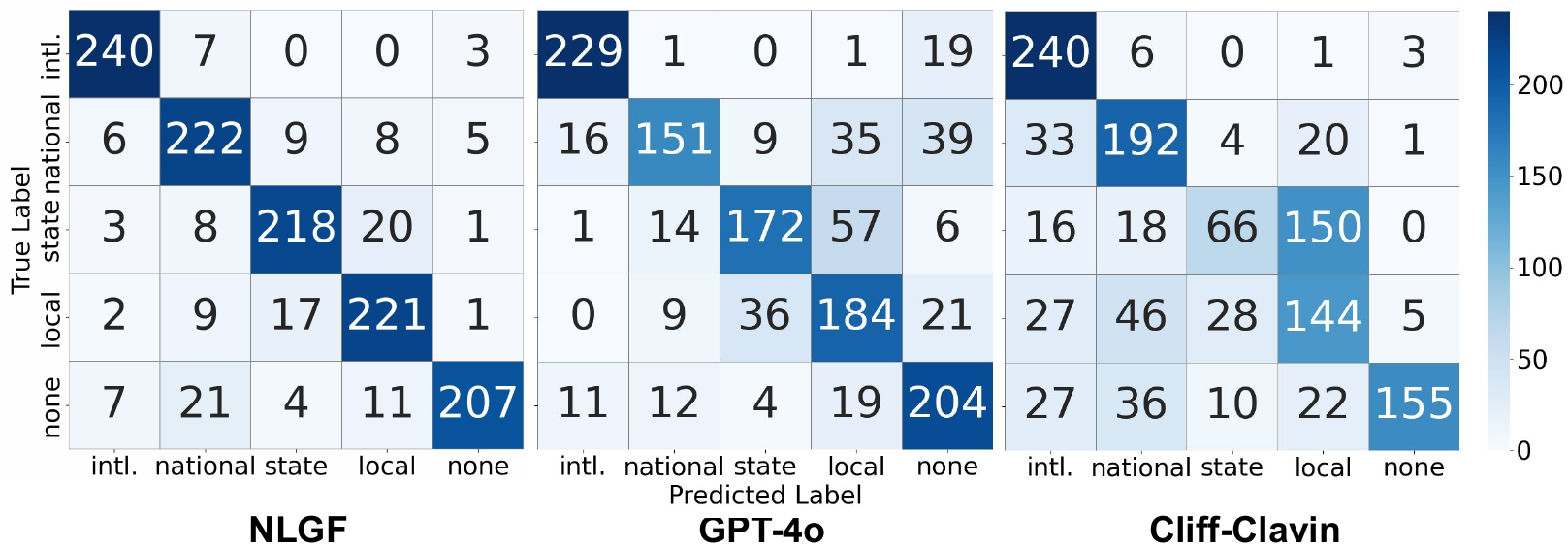}}
\caption{NLGF vs. GPT-4o vs. Cliff-Clavin: Geo-Focus Level Classification Confusion Matrices.}
\label{fig:xgb_gpt}
\Description{NLGF vs. GPT-4o vs. Cliff-Clavin: Geo-Focus Level Classification Confusion Matrices.}
\end{figure}

Fig.~\ref{fig:xgb_gpt} presents the confusion matrices of all three models. Across all geo-focus levels, NLGF achieved higher $F_1$ than GPT-4o and Cliff-Clavin, reflecting its stronger ability to capture spatial cues and interpret geographic information embedded in news text. All models showed their strongest performance on the \textit{international} class, which is unsurprising: international news typically contains globally identifiable entities and unambiguous geopolitical markers, making them easier to distinguish from domestic ones. The \textit{none} class also exhibits relatively strong performance for NLGF, indicating that the model can effectively identify articles lacking a clear geographic focus. In contrast, there was more confusion between \textit{local}, \textit{state}, and \textit{national}. This is unsurprising, since local news stories could include broader regional or national news. Similarly, state news reporting can easily be conflated with local news. Also, national news sometimes adopts a localized framing to increase relevance, leading to misclassifications. These confusions, rather than indicating model failure, reflect the inherent complexity and semantic overlap of geographic references in news. GPT-4o performed poorly in classifying the \textit{state} and \textit{national} classes, where spatial cues such as the publisher's location are essential. Unlike GPT-4o, which lacks persistent spatial grounding, the NLGF model leverages explicit spatial indicators, resulting in significantly fewer misclassifications in boundary cases. Cliff-Clavin exhibits notable weakness in distinguishing the \textit{local} and \textit{state} geo-focus classes, highlighting the limitation of frequency-based methods. In contrast, NLGF accounts for spatial hierarchy and the contextual prominence of toponyms, allowing it to more reliably distinguish domestic geo-focus levels and handle articles with overlapping geo-focus signals.

NLGF also outperformed GPT-4o and Cliff-Clavin~\cite{d2014cliff} in geo-foci identification (Table~\ref{tab:gpt_vs_xgb_cliff}), achieving a precision, recall, and $F_1$ of 0.86, 0.89, and 0.86, respectively, compared to GPT-4o (0.65, 0.69, and 0.66) and Cliff-Clavin (0.29, 0.64, 0.37). These findings further demonstrate that NLGF’s explicit integration of geographic disambiguation with spatially informed classification yields significant improvements over both LLM-only and gazetteer-only approaches.

Collectively, the results validate the NLGF's hybrid architecture: leveraging LLM for toponym disambiguation with spatial-semantic feature representation for classification, and heuristic scoring, are effective for geo-focus level/geo-foci prediction for US local news.

\section{Conclusions and Future Work}

Does local journalism address the information needs of the communities they serve? To help address this question, we focus specifically on identifying the geographic regions covered by US local news (geo-foci) with NLGF. NLGF detects the most salient administrative division (geo-focus level) and the most salient geographic locations (geo-foci) of US local news articles. It ($F_1$: 0.86) significantly outperformed GPT-4o ($F_1$: 0.66) and Cliff-Clavin ($F_1$: 0.37) by combining LLMs for toponym disambiguation with spatial-semantic features for geo-focus level classification, and heuristic scoring for geo-foci identification.

We propose these improvements to NLGF in future work. First, NLGF assigns a single geo-focus level, even though local news article can have multiple geo-focus levels (e.g., \textit{local} and \textit{national}). We propose to implement a multi-label geo-focus level classifier. Second, while this study focused on US news, the framework is readily adaptable to other countries, and future research will investigate its applicability to other countries.

NLGF is a valuable contribution to the challenge of geo-focus identification in local news, with broad implications for computational journalism, particularly in measuring the nationalization of local news where NLGF can be applied to assess shifts from local to national narratives. In short, NLGF could enable researchers to better study local media.

\begin{acks}
We thank the NSF for funding this work (award no. 2245508). The NSF had no role in designing our study, data collection and analysis, decision to publish, or preparation of the manuscript. The authors acknowledge William \& Mary Research Computing for providing computational resources and technical support that have contributed to the results reported within this article.
\end{acks}

\clearpage

\printbibliography

\end{document}